\documentclass{article}
\usepackage{spconf,amsmath,amssymb,graphicx}
\usepackage{subcaption}

\title{MULTIMODAL PUNCTUATION PREDICTION WITH CONTEXTUAL DROPOUT}
\name{Andrew Silva$^{\star}$ \thanks{$^{\star}$ Work done during an internship at Apple.} \qquad Barry-John Theobald$^{\dagger}$ \qquad Nicholas Apostoloff$^{\dagger}$}
  
  \address{$^{\star}$ Georgia Institute of Technology $\qquad$
      $^{\dagger}$ Apple Inc.}

\begin{document}
%\ninept
%
\maketitle
\begin{abstract}
Automatic speech recognition (ASR) is widely used in consumer electronics. ASR greatly improves the utility and accessibility of technology, but usually the output is only word sequences without punctuation. This can result in ambiguity in inferring user-intent.  We first present a transformer-based approach for punctuation prediction that achieves 8\% improvement on the IWSLT 2012 TED Task, beating the previous state of the art \cite{courtland-etal-2020-efficient}.  We next describe our multimodal model that learns from both text and audio, which achieves 8\% improvement over the text-only algorithm on an internal dataset for which we have both the audio and transcriptions. Finally, we present an approach to learning a model using contextual dropout that allows us to handle variable amounts of future context at test time. 
\end{abstract}
\section{Introduction}
\label{sec:intro}

Speech is a primary source of input for many digital assistants, where an automatic speech recognizer (ASR) transcribes speech for downstream language processing.  However, ASR is concerned with transcribing \emph{what}, rather than \emph{how}, something was said.  Thus transcriptions typically are sequences of words without structure, which can result in ambiguity in resolving user-intent.  For example, consider the sequence of words ``text mom and dad will be there soon.''  A digital assistant might interpret this word sequence as a command to send a text message to the recipients ``mom'' and ``dad'' with the message ``will be there soon.''  However, if the word sequence included punctuation, e.g.\ ``text mom, and dad will be there soon'' then an alternative interpretation might be to send a text message to the recipient ``mom'' containing the message ``and dad will be there soon.'' 

Our primary contributions are: (1)  multimodal punctuation prediction using text features from a transformer model fused with learned audio features to consider speaking style to improve prediction accuracy.  (2) An investigation of the effects of future context on punctuation prediction performance, and a training scheme (contextual dropout) to improve performance with varying amounts of future information.
%An investigation of the effects of future context on the performance of punctuation prediction systems, and a training scheme (contextual dropout) to improve performance with varying amounts of future information.

\section{Related Work}
\label{sec:related-work}

Early approaches for punctuation prediction used decision trees, multi-layer perceptrons (MLPs), hidden Markov models (HMMs), or finite-state models trained using n-grams or prosodic features \cite{christensen2001punctuation}. However, the complexity of punctuation prediction required simplifying assumptions about the symbols to consider, or reduced the problem to sentence-boundary detection \cite{wang2012dynamic}. More powerful models have included conditional random fields \cite{lu-ng-2010-better}, boosted hierarchical prediction \cite{kolavr2012development}, and punctuation as neural machine translation (NMT) \cite{peitz2011modeling}. However, state of the art results have been achieved using convolutional neural networks (CNNs) \cite{zelasko2018punctuation}, long short term memory (LSTM) networks \cite{xu2016lstm,tilk2015lstm,klejch2017sequence}, and transformers \cite{courtland-etal-2020-efficient,vaswani2017attention}, treating the problem as a classification task. In these latter approaches, the task is to consider which punctuation symbol should follow each token in an utterance, rather than, say, detecting just sentence boundaries in text.

Various datasets have been proposed for punctuation prediction, including the Fisher corpus \cite{zelasko2018punctuation}, TED talks and journalism data \cite{ueffing2013improved}, and the IWSLT 2012 TED task \cite{courtland-etal-2020-efficient,tilk2015lstm,che2016punctuation}. We evaluate our unimodal text-based approach on the IWSLT TED Task dataset, comparing a baseline version of our approach to prior CNN, LSTM, and transformer architectures.  We then evaluate a multimodal approach on an internal dataset for which we have the speech and accompanying ground-truth transcriptions.

Transformers have been applied to punctuation prediction. For example, using Word2Vec \cite{mikolov2013distributed} and Speech2Vec embeddings \cite{chung2018speech2vec}, Yi and Tao \cite{yi2019self} train an attention model to classify punctuation in sequences. Courtland et al.\ \cite{courtland-etal-2020-efficient} apply a RoBERTa model to punctuation prediction on the IWSLT 2012 TED Task, using 100-token sequences for prediction and aggregating multiple predictions for each token, creating a pseudo-ensemble of transformers with varying contexts. We instead use a single prediction for each token, and we find that we can achieve superior performance using much smaller context windows than \cite{courtland-etal-2020-efficient}. Finally, \cite{sunkara-etal-2020-robust,sunkara2020multimodal} apply transformers to punctuation prediction using lexical features and prosodic features which are aligned using pre-trained feature extractors and alignment networks. In contrast to \cite{sunkara-etal-2020-robust,sunkara2020multimodal}, we use forced-alignment from ASR and learn acoustic features from scratch from spectrogram segments corresponding each text tokens.

\section{Approach}
\label{sec:approach}
\begin{figure}[t]
\centering
        \includegraphics[width=0.9\linewidth]{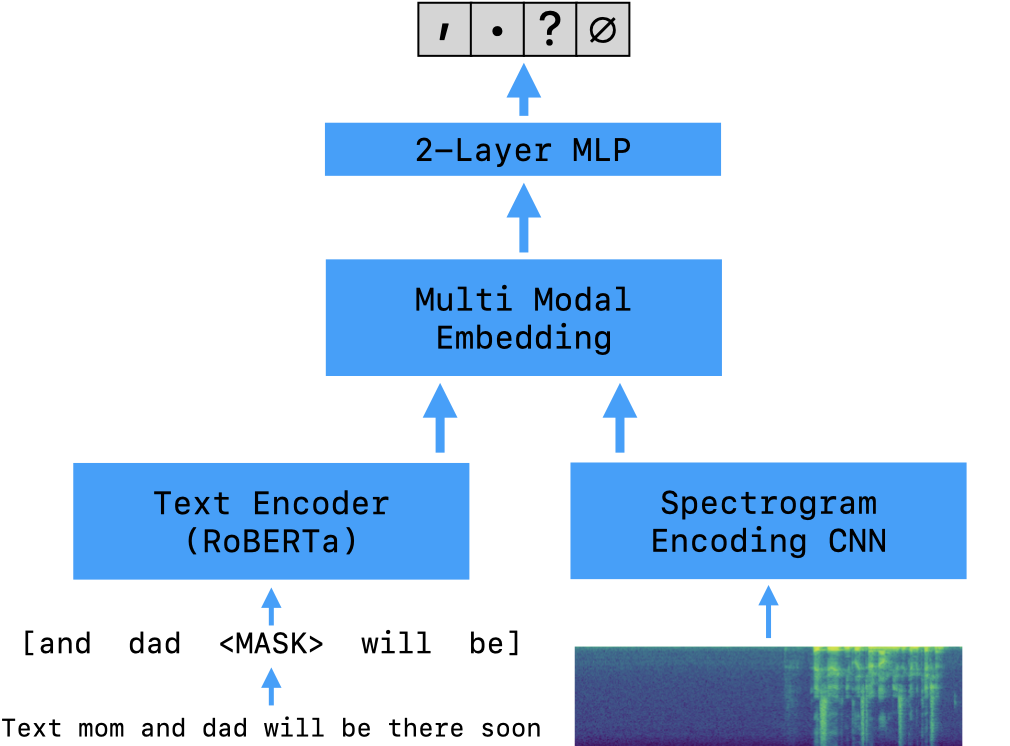}
        \caption{The architecture of our multimodal punctuation prediction model with training context width $C_T=2$.}
        \label{fig:model-architecture}
\end{figure}

Our architecture for punctuation prediction using both text and audio features is shown in Figure \ref{fig:model-architecture}. In this work, we consider the punctuation symbols in the set \{`,', `.', `?', `$\varnothing$'\}, where $\varnothing$ indicates no punctuation symbol. We select this particular set of punctuation symbols as these are best represented in the datasets that we use.  As we are interested in improving the structure of ASR transcriptions, we consider input data which has been force-aligned with audio, allowing us to easily retrieve clips of audio relevant to each recognized token in the utterance.

Transcribed utterances are tokenized using a pre-trained RoBERTa tokenizer \cite{liu2019roberta,Wolf2019HuggingFacesTS}. For each token, $u_t$, we construct a contextually-relevant sequence of a pre-specified length. For example, given a context window of two past and two future tokens, the input sequence:
\begin{displaymath}
[u_{t-2}, u_{t-1}, u_t, \langle MASK \rangle, u_{t+1}, u_{t+2}],
\end{displaymath}
is input to the model with the objective, replace $\langle MASK \rangle$ with the correct punctuation symbol. Missing information at the beginning and end of a transcription is replaced with $\langle PAD \rangle$ tokens.

Text-based features are extracted by fine-tuning a pre-trained RoBERTa masked language model \cite{liu2019roberta,Wolf2019HuggingFacesTS}. The context window for a given token is passed to the model and all hidden layers corresponding to $\langle MASK \rangle$ are concatenated to form a 9984-D feature vector.

Acoustic features for a given token, $u_t$, are extracted by first using the ASR alignment to identify the mid-point of the token in the audio, and then extracting a window of size $2\times C^A$ centered on $u_t$, where $C^A$ is the context window size for acoustic data.  Note that the context width of the text and the context width of the audio need not align.  We experimented with different values of $C^A$ and found that $C^A=1$ second worked well compared with longer audio context widths.  Spectrogram features are extracted from the resulting audio, which are encoded by a CNN into a 4096 dimension vector.  This latent representation for the audio is then concatenated with the text-based features, and these are subsequently passed through two fully-connected layers to predict the most likely punctuation class for $\langle MASK \rangle$.

For improving the robustness of punctuation prediction, we experiment both with dropping \emph{variable future contextual information}, and substituting word tokens.  In particular, 15\% of randomly selected future input tokens are changed to $\langle$DROP$\rangle$, 15\% of the inputs samples have half the future context removed, 1.5\% of the samples have no future context, and 1.5\% of the tokens are swapped for other word tokens selected randomly \cite{devlin2018bert}.  We do not predict punctuation classes or ablated tokens for the substituted $\langle$DROP$\rangle$ inputs; rather, they serve to help the model generalize to noisy or missing future context, as would be expected in a deployed ASR system.

In all of our experiments, we use an audio context window $C^A=1$s, and we establish the upper bound of useful prior text context at 32 tokens, beyond which we saw little improvement in accuracy. Both the audio and the text context windows were set using an empirical hyper-parameter search, though we note that future work could explore automated discovery of the appropriate amount of context, or the use of dynamic context for different domains (e.g. very little context for live transcripts and more context for pre-recorded transcripts). We also train all of our models using balanced class sampling after experiments showed that the inherent class imbalance in the data adversely affected final task performance.

\section{TED Task}
\label{sec:ted-task}
\begin{table}
\centering
\begin{tabular}{ lccccc }
 \hline
    Dataset & `,' & `.' & `?' & `$\varnothing$' & Total \\

 \hline
 IWSLT 2012 TED \cite{che2016punctuation} & 830 & 808 & 46 & 10942 & 12626\\
 Our re-tokenized version & 718 & 797 & 46 & 13579 & 15140
\end{tabular}
\caption{A comparison of dataset statistics for the IWSLT 2012 TED data and our internal dataset.}
\label{table:dataset-stats}
\end{table}
We first benchmark on the IWSLT 2012 TED task data \cite{iwslt2012data}, which has been used previously for punctuation prediction from raw text \cite{courtland-etal-2020-efficient,che2016punctuation}. Prior work uses a standardized train and test set, tokenized using the GloVe tokenizer \cite{pennington2014glove}. As our work is built on the RoBERTa encoder from Hugging Face \cite{Wolf2019HuggingFacesTS}, tokenization with the standard GloVe tokenizer is not appropriate. We therefore recreate the dataset using the RoBERTa tokenizer, resulting in dataset statistics that differ from prior work. We include a comparison in Table \ref{table:dataset-stats}, where we can see that word-piece tokenization results in significantly more $\varnothing$ tokens. In the interest of comparing to prior work \cite{tilk2015lstm,che2016punctuation}, we train an LSTM predictor and a CNN predictor on our re-tokenized dataset and include their performance in our results. Our hyper-parameter search was conducted using 85\% of the training dataset, with the remaining 15\% as a validation set, though we used all available training data for the final model.

\begin{table}
\centering
\begin{tabular}{ lccccc }

   $C_T =$  & $0$ & $2$ & $8$ & $32$ & 32 - with \\
   & & & & & dropout\\
 \hline
 RoBERTa & & & & & \\
 
  $C_E=0$ & \textbf{57.7} & 50.3 & 45.7 & 40.3 & 53.2\\
  $C_E=32$ & 58.6 & 81.3 & \textbf{84.1} & 82.8 & 83.2\\
 LSTM & & & & & \\
 $C_E=0$ & \textbf{46.6} & 36.4 & 36.8 & 39.3 & 45.4\\
 $C_E=32$ & 44.2 & 69.2 & 70.5 & \textbf{72.2} & 71.6\\
 CNN & & & & & \\
 $C_E=0$ & \textbf{38.7} & 34.5 & 31.1 & 34.5 & 38.0\\
 $C_E=32$ & 28.1 & 24.0 & 52.6 & \textbf{63.5} & 57.7\\
 \hline
\end{tabular}
\caption{Mean F1 scores across a range of training contexts but tested with future context sizes of 0 and 32. Our contextual dropout provides a good balance for accuracy with no future context and full future context. }
\label{table:ted-results}
\end{table}

\begin{figure*}[t]
\centering
\begin{subfigure}[b]{\textwidth}
    \centering
    \includegraphics[width=0.6\textwidth]{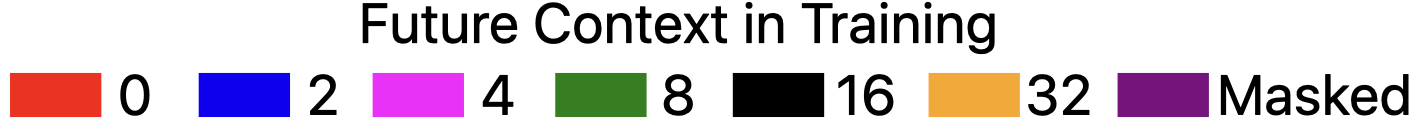}
\end{subfigure} 
    \begin{subfigure}[b]{0.25\textwidth}
        \includegraphics[width=\textwidth]{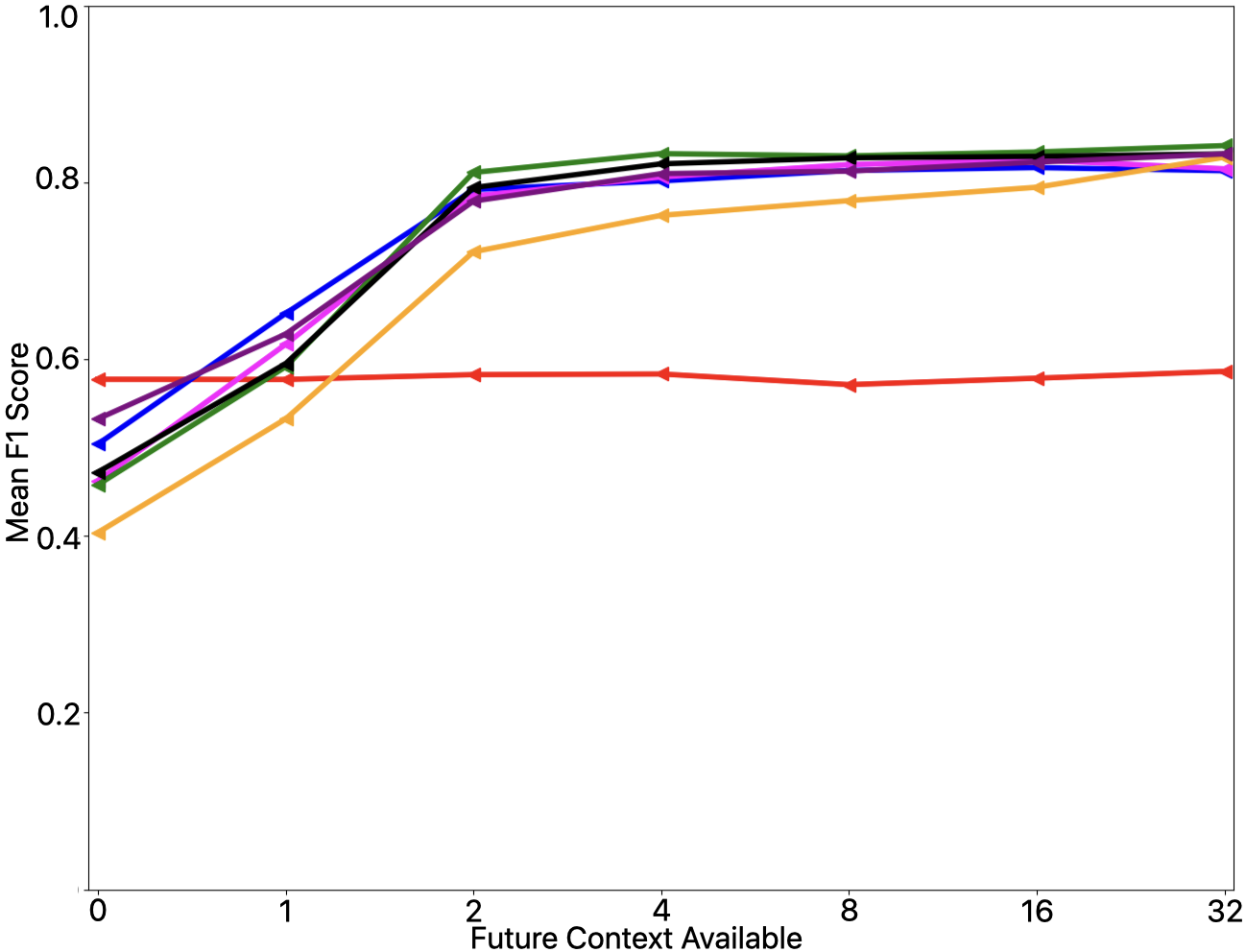}
        \caption{RoBERTa}
        \label{fig:roberta-ted-results}
    \end{subfigure}
    ~~
    \begin{subfigure}[b]{0.25\textwidth}
        \includegraphics[width=\textwidth]{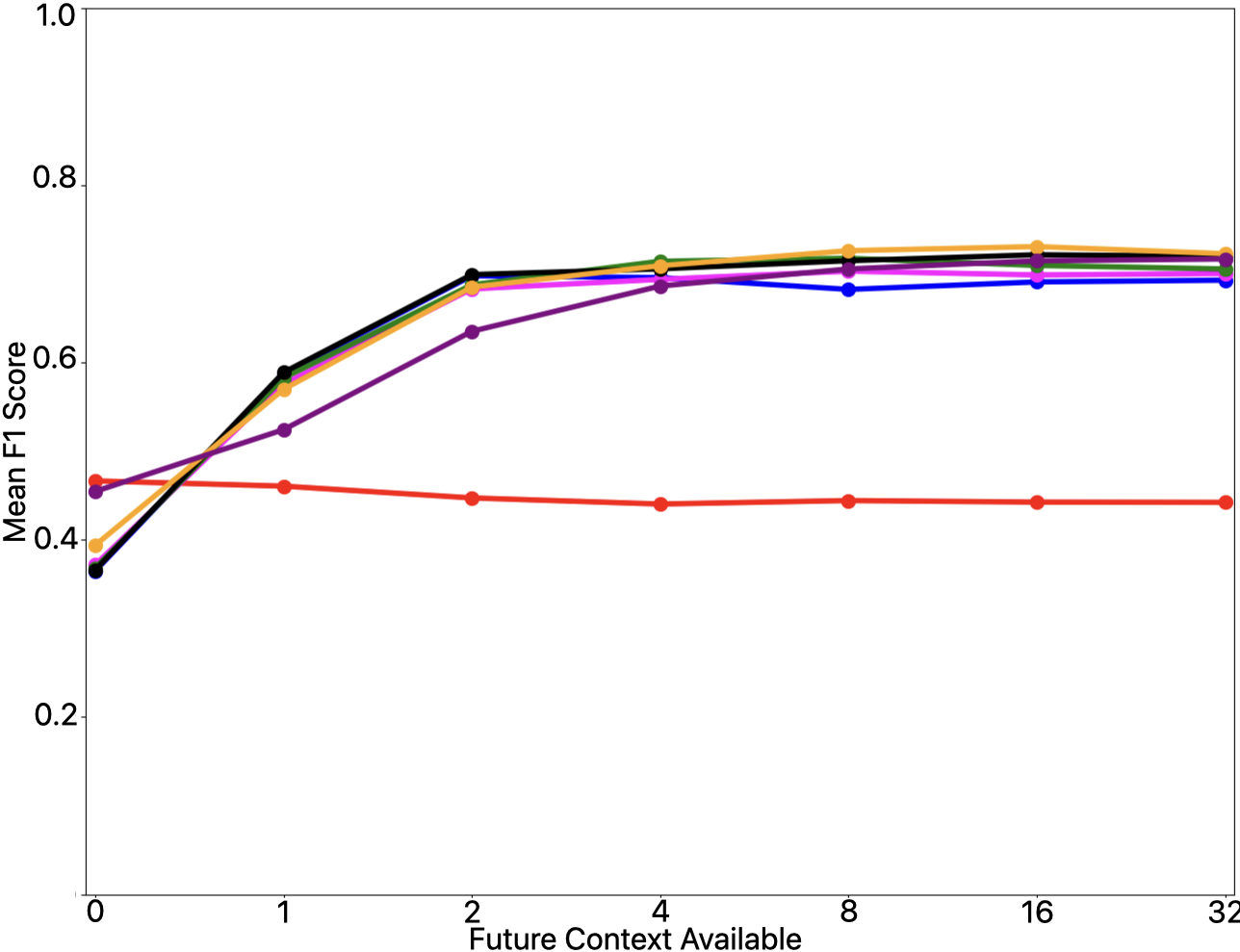}
        \caption{LSTM}
        \label{fig:lstm-ted-results}
    \end{subfigure}
    ~~
    \begin{subfigure}[b]{0.25\textwidth}
        \includegraphics[width=\textwidth]{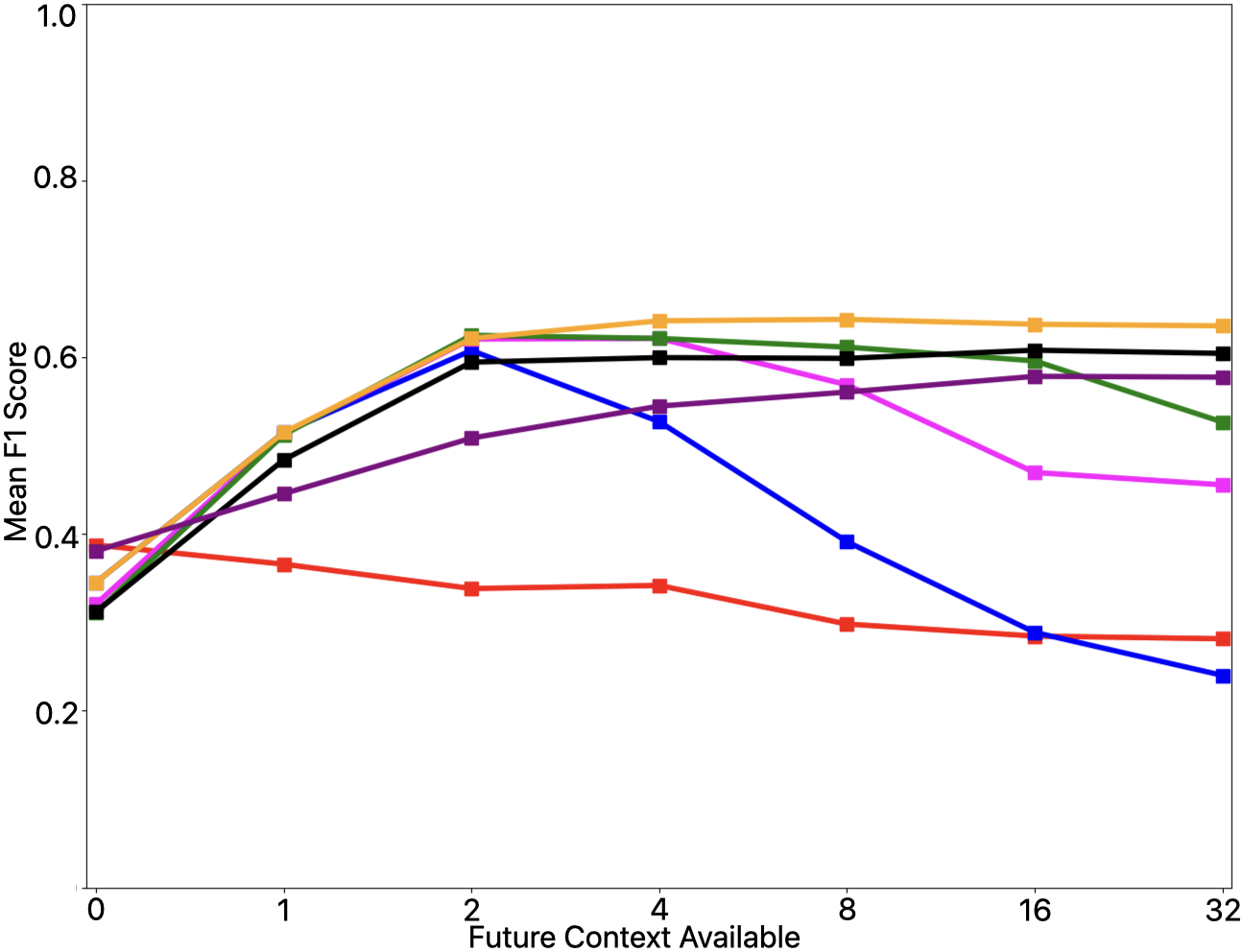}
        \caption{CNN}
        \label{fig:cnn-ted-results}
    \end{subfigure}
\caption{A comparison of Mean F1 scores as a function of available future context in test data for RoBERTa, LSTM, and CNN models. Each line indicates the amount of future context available during training for $C_T=0$ (red), $C_T=2$ (blue), $C_T=4$ (pink), $C_T=8$ (green), $C_T=16$ (black), $C_T=32$ (yellow), and $C_T=32$ with contextual dropout (purple).}
\label{fig:ted-results}
\end{figure*}

We evaluate the impact of our masking approach on robustness to \emph{variable future information} for the transformer, LSTM, and CNN models. We train models with varying amounts of available future context, $C_T$, in the range $C_T \in [0,2,4,8,16,32]$ tokens, and we also train a model that uses $C_T=32$ with randomly selected tokens in the future context dropped. We then evaluate each model with future context, $C_E$, in the range $C_E \in [0,2,4,8,16,32]$, allowing us to see how well each model generalizes to different amounts of available future context.  For example, training with no future context, $C_T=0$, and testing with each of $C_E \in [0,2,4,8,16,32]$.  We expect performance to be maximized when $C_T$ is equal to $C_E$, but that randomly dropping tokens in training provides the best balance of performance for all values of $C_E$. All models, regardless of the setting of $C_T$, have 32 tokens of past context available during training and testing. The results from this experiment are presented in Figure \ref{fig:ted-results} and Table \ref{table:ted-results}.

Unsurprisingly, we find that more context generally yields higher performing models, with $C_T=32$ providing the strongest LSTM and CNN models. However, we observe that the effects of increased context are less significant than we expected, and after $C_T=8$, the performance improvement is negligible. Furthermore, we observe that performance is not necessarily maximized when $C_T$ and $C_E$ are the same. Rather, higher values of $C_E$ generally correspond to better performance across the board, regardless of the value of $C_T$. This finding indicates that the models are able to use greater contextual information as it is available, but that they have not overfit to the exact context with which they were trained. The notable exception to this trend is the CNN architecture, which routinely shows drops in performance wherever $C_T$ and $C_E$ are not equal. 

Our TED Task experiment confirms that the transformer is the best model for punctuation prediction, far exceeding the performance of the LSTM and CNN. We also show that training with a large context window, as in \cite{courtland-etal-2020-efficient}, is unnecessary, as setting $C_T=8$ yields similar performance as setting $C_T=32$. Finally, we show that training with a context-size of 32, but including context-dropout provides robustness to missing context at test time since the model is able to better generalize to unseen context window sizes.

\section{Podcast Task}
We use a large corpus of internal data to evaluate the effects of available future context on performance, the impact of context-dropout on generalization, and to measure the impact of adding audio features from speech. This dataset consists of approximately 1400 hours of manually transcribed speech from podcasts and from an internal developer talk dataset. We test on approximately 340 hours of podcast audio for which manual transcriptions are available.
%We note that the training data includes a subset of the podcasts in the testing data, but that the specific episodes are distinct between the datasets.  
The test set includes podcasts from the training set, but with unseen episodes, and a broader set of entirely unseen podcasts, and so the model must generalize to unseen speakers and different discussion formats.

\begin{table}
\centering
\begin{tabular}{ lcccc }
     & (A) & (B) & (C) & (D) \\
     \hline
RoBERTa & & & & \\
 
  $C_E=0$ & 36.6 & 40.1 & 40.6 & \textbf{55.4}\\
  $C_E=32$ & 73.8 & 70.6 & \textbf{81.5} & 78.9\\
 LSTM & & & & \\
 $C_E=0$ & 32.9 & 38.2 & 36.4 & \textbf{46.7} \\
 $C_E=32$ & 66.8 & 64.1 & \textbf{69.3} & 66.9 \\
 CNN & & & & \\
 $C_E=0$ & 34.5 & 32.6 & 41.0 & \textbf{43.1} \\
 $C_E=32$ & 58.2 & 53.0 & \textbf{60.0} & 58.5\\
 \hline
\end{tabular}
\caption{Mean F1 scores for:  text-only models (A and B), and multimodal models (C and D).  Models (A and C) were trained without dropout, and models (B and D) were trained with dropout.  Audio improves accuracy for all base models, where the mean F1 score is increased by up to 10\%.  Context-dropout and adding audio together always improve performance when no future information is available.}
\label{table:internal-results}
\end{table}

We seek to understand the effects of a multimodal approach on punctuation prediction with varying amounts of future information. While multimodal approaches are common for punctuation prediction \cite{klejch2017sequence,sunkara2020multimodal,yi2019self}, we are the first to incorporate learned acoustic features from scratch using force-alignment from ASR rather than relying on other data to pre-train or hand-select acoustic features.

We performed a sweep over context window sizes for the TED Task to identify the best performing context size for each of the transformer, LSTM, and CNN.  The chosen context size for a particular model is then used to train a model on our internal dataset, where we train four versions of each model type: text-only no dropout, multimodal (text and audio) no dropout, text-only with dropout, and multimodal with dropout. Training these four variants allows us to see the effects of adding audio and/or training with context-dropout. We note that the context-dropout applies only to the text input and that augmentations of the spectrogram, as in \cite{jaitly2011new}, are left to future work. When we evaluate each model on the test set, we again evaluate on the test set with $C_E=0$ and $C_E=32$. Results for all approaches are given in Table \ref{table:internal-results}.

As shown in Table \ref{table:internal-results}, the mean F1 score increases as more future context is available at test time for all unimodal (rows per model in columns A and B) and multimodal models (rows per model in columns C and D).  Also, all models achieve a significant gain in accuracy after adding audio (compare columns A and C, and columns B and D).  We find that after training with contextual dropout, the models perform significantly better when no future context is available at test time (compare columns A and D in rows marked $C_E = 0$).

\begin{figure}[t]
\centering
\begin{subfigure}[b]{0.2\textwidth}
        \includegraphics[width=\textwidth]{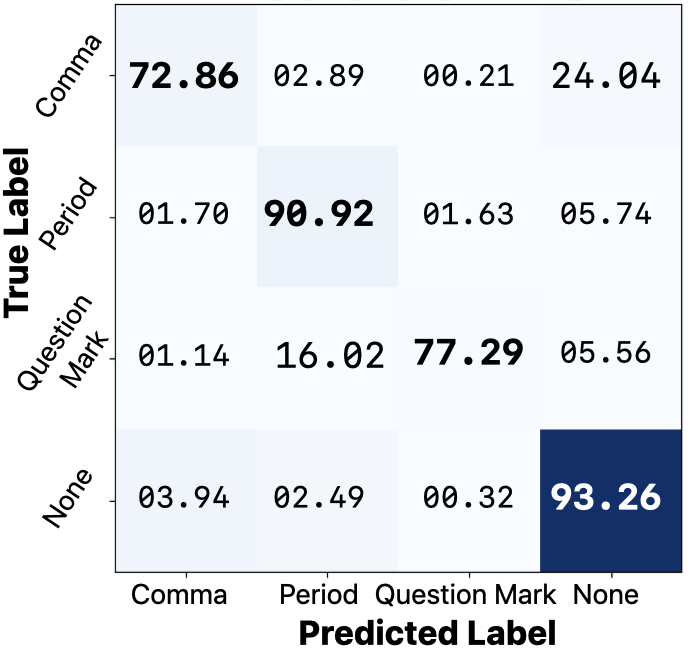}
        \caption{Without Audio}
        \label{fig:internal-cm-text}
    \end{subfigure}
    ~~
    \begin{subfigure}[b]{0.2\textwidth}
        \includegraphics[width=\textwidth]{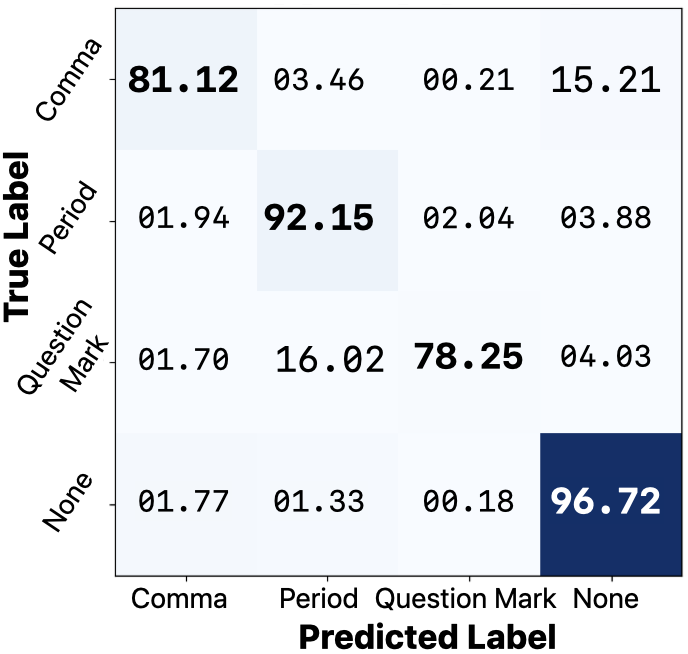}
        \caption{With Audio}
        \label{fig:internal-cm-audio}
    \end{subfigure}
\caption{Confusion matrices for a RoBERTa text encoder with $C_T=32$ and with/without 1 second spectrograms for audio, evaluated with $C_E=32$. Even an imbalance towards None, the models achieve high accuracy on `.', and do well on the remaining classes. Audio improves `,' accuracy by 9\%. The brightness of a cell relates to the relative number of samples for that class, where darker squares indicate more samples.}
\label{fig:internal-cm}
\end{figure}

For a more granular analysis of our best-performing model (RoBERTa), Figure \ref{fig:internal-cm} shows the confusion-matrices before and after adding audio for $C_T=C_E=32$.  We can see that across the board all symbols are better predicted using the multimodal vs.\ the text-only model.  Of particular note, we see that the text-only model often mistakes commas for $\varnothing$, but a significant number of these are corrected with the integration of audio.  This observation suggests the model better recognizes when a short pause is a natural break in speech.  
%We also observe that audio data does not resolve confusion between `.' and `?'. This implies that the model is accurate for identifying endpoints in sequences, but not at resolving \textit{which} endpoint is used. An analysis into specific examples which were ``confused'' did not reveal any common patterns, and we speculate that the errors may be attributed to the podcast data rather than more natural face-to-face interaction.

\section{Conclusion}
\label{sec:conclusion}
We have described a multimodal approach for punctuation prediction that uses a RoBERTa encoder for text and a spectrogram encoder for speech.  Our system outperforms previous approaches, and it does not require pre-trained lexical-audio alignment networks.  Furthermore, we show that punctuation prediction is less future-dependent than suggested in prior work, where only one or two tokens of posterior context is required for high accuracy. Finally, we proposed context-dropout in training to improve robustness and generalization.

Future work will investigate an iterative decoding scheme that leverages past predictions to inform future classifications, considering the overall likelihood of different sets of punctuation placements for a complete sequence. We will also consider the problem of punctuating text in languages other than English, and consider a broader set of symbols than the four considered in this work.

\section{Acknowledgements}
We are grateful to Sue Tranter and Tali Singer for assistance with data preparation, and to Russ Webb, Ashish Shrivastava and Ahmed Hussen Abdelaziz for valuable feedback.

\bibliographystyle{IEEEbib}
\bibliography{refs}

\end{document}